%% file: paper.tex
\documentclass[runningheads]{llncs}
\usepackage[T1]{fontenc}
\usepackage{graphicx}
\usepackage{bbding}
\usepackage[section]{placeins}
\usepackage{booktabs}

\usepackage{tikz}
\usepackage{amsmath,amssymb}
\usetikzlibrary{
  arrows.meta,
  positioning,
  calc,
  fit,
  backgrounds,
  decorations.pathreplacing,
  decorations.markings,
  shapes.geometric,
  shapes.misc,
  patterns,
  shadows,
  fadings
}

\definecolor{vaecol}{HTML}{4A90D9}
\definecolor{unetcol}{HTML}{E67E22}
\definecolor{condcol}{HTML}{27AE60}
\definecolor{disent}{HTML}{8E44AD}
\definecolor{timecol}{HTML}{C0392B}
\definecolor{noisecol}{HTML}{7F8C8D}
\definecolor{skipcol}{HTML}{16A085}
\definecolor{supercol}{HTML}{D4AC0D}
\definecolor{vaelight}{HTML}{D6EAF8}
\definecolor{unetlight}{HTML}{FDEBD0}
\definecolor{disentlight}{HTML}{E8DAEF}

\definecolor{dig2col}{HTML}{2980B9}
\definecolor{dig3col}{HTML}{27AE60}
\definecolor{dig5col}{HTML}{D35400}
\definecolor{dig8col}{HTML}{8E44AD}
\definecolor{oodlight}{HTML}{FEF9E7}

\definecolor{jointBlue}{HTML}{2E5EAA}
\definecolor{decompOrange}{HTML}{E8762B}
\definecolor{bifYellow}{HTML}{FFF3C4}
\definecolor{bifBorder}{HTML}{D4A017}
\definecolor{noiseGray}{HTML}{8B8B8B}
\definecolor{dataGreen}{HTML}{2D8E4E}
\definecolor{nullPurple}{HTML}{7B4F9D}
\definecolor{bgLight}{HTML}{F7F9FC}
\definecolor{cfmTeal}{HTML}{1A8A7D}
\definecolor{ddimIndigo}{HTML}{4B3F8F}
\definecolor{accentRed}{HTML}{CC3344}

\tikzset{
  blk/.style={
    draw, rounded corners=2pt, minimum height=0.48cm,
    minimum width=1.3cm, align=center, font=\tiny\sffamily, thick
  },
  convb/.style={blk, fill=vaecol!15, draw=vaecol!70},
  resb/.style={blk, fill=vaecol!25, draw=vaecol},
  cresb/.style={blk, fill=unetcol!20, draw=unetcol},
  embb/.style={blk, fill=disent!20, draw=disent},
  timeb/.style={blk, fill=timecol!15, draw=timecol!70},
  opb/.style={blk, fill=noisecol!10, draw=noisecol!60},
  outb/.style={blk, fill=condcol!12, draw=condcol!70},
  dim/.style={font=\tiny\ttfamily, text=black!55},
  sublbl/.style={font=\tiny\sffamily},
  arr/.style={-{Stealth[length=3.5pt]}, semithick},
  darr/.style={-{Stealth[length=3pt]}, semithick, dashed},
  sarr/.style={-{Stealth[length=3pt]}, semithick, skipcol, densely dashed},
}

\usepackage{algorithm}
\usepackage{algpseudocode}

\begin{document}
\title{OOD Generalization as a Bifurcation Problem}
\author{Nguyen-Thanh-Luong Doan\inst{1},
Quang-Vu Nguyen\inst{1}\textsuperscript{,\Envelope}\thanks{\Envelope~Corresponding author}, Tang-Phu-Quy Le\inst{1}, and Cong-Phap Huynh\inst{1}}
\authorrunning{Nguyen-Thanh-Luong et al.}
\institute{The University of Danang - Vietnam-Korea University of Information and Communication Technology \\
\email{\{dntluong, nqvu,  quyltp.22git, hcphap\}@vku.udn.vn}}
\maketitle              % typeset the header of the contribution

% \begin{figure}
% \includegraphics[width=\textwidth]{images/plot_interference.png}
% \caption{\lp{need to have a caption}}
% \end{figure}

%
\begin{abstract}
Systematic out-of-distribution (OOD) generation remains a critical bottleneck for continuous-time generative models. While standard joint classifier-free guidance (CFG) routinely fails to synthesize unobserved concept combinations, exact decomposed scoring generalizes robustly at the cost of severe computational overhead. In this work, we reveal that compositional binding is not a uniform process but a highly localized phase transition. We identify the ``semantic bifurcation window''---the precise temporal interval where joint and decomposed vector fields meaningfully diverge. Exploiting this dynamic, we propose \textbf{surgical guidance}, a hybrid sampling strategy that restricts exact multi-pass scoring strictly to this critical window. On an OOD bi-digit MNIST testbed, surgical guidance achieves state-of-the-art compositional fidelity at a fraction of the inference cost, yielding a $+5.3\%$ absolute improvement in pairwise accuracy over the joint baseline by intervening during just the first $15\%$ of the diffusion trajectory. Furthermore, our empirical analysis uncovers a fundamental topological divide: diffusion models (SDEs) force conceptual resolution immediately at peak noise, whereas Conditional Flow Matching (ODEs) delays structural binding until intermediate features emerge, establishing a new temporal framework for accelerating large-scale generative decoding.
\end{abstract}
\input{intro}

\input{background}

\input{bifurcation_window}

\input{experiments}

\input{conclusion}

\input{ack}

\bibliographystyle{splncs04}
\bibliography{refs}

\end{document}

%% file: intro.tex
\section{Introduction}

Systematic compositional generalization is a fundamental challenge for deep generative models \cite{chomsky1965aspects,higgins2017betavae,marcus2022very,russin2024frege}. While continuous-time models like diffusion and Conditional Flow Matching (CFM) \cite{lipman2023flow} exhibit remarkable synthesis capabilities, they consistently struggle to generate out-of-distribution (OOD) concept combinations unseen during training. Standard joint classifier-free guidance (CFG) \cite{ho2021classifierfree} models in-distribution data well but fails to disentangle and recombine independent features. Conversely, decomposed scoring paradigms \cite{du2020compositional,liu2022compositional}---rooted in the product-of-experts formulation of Energy-Based Models and Composable Diffusion---achieve robust OOD generalization. However, this explicit mathematical factorization demands multiple forward network passes per integration step, introducing severe computational overhead.

\begin{figure}[!htbp]
    \centering
    \input{images/fig_ood_mnist}
    \caption{Bi-digit MNIST out-of-distribution (OOD) generalization testbed.}
    \label{fig:ood_mnist}
\end{figure}
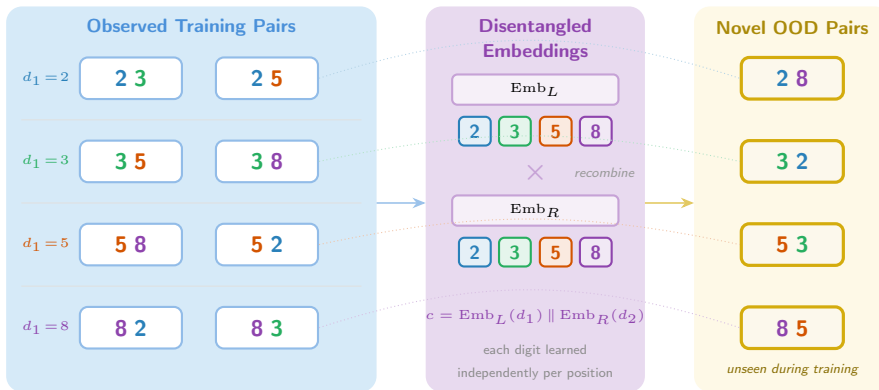

To systematically investigate this inference trade-off, we establish a controlled bi-digit MNIST \cite{lecun2010mnist} testbed. We train models to generate paired digits from a heavily restricted subset of observed combinations. True generalization requires the model to correctly synthesize novel, unobserved digit pairs strictly by recombining disentangled left and right positional embeddings (Figure \ref{fig:ood_mnist}). This isolated environment allows us to track the exact dynamics of compositional binding.

By analyzing the vector fields of both SDE (Diffusion) and ODE (CFM) trajectories, we demonstrate that joint and decomposed scores do not conflict uniformly across time. Instead, they exhibit a localized phase transition. We formalize this temporal region as the \textit{semantic bifurcation window}, defining the precise interval where the generative process actively resolves compositional conflicts. 

Exploiting this dynamical behavior, we propose \textbf{surgical guidance}. This dynamic sampling strategy dynamically switches objective functions during generation, applying exact decomposed scoring strictly within the bifurcation window while utilizing computationally efficient joint CFG everywhere else. 

Surgical guidance matches the OOD performance of full decomposed generation at a fraction of the inference cost. Furthermore, our empirical evaluations expose a distinct topological divergence between generative frameworks: diffusion models force compositional binding almost immediately at peak noise, whereas flow matching delays structural resolution until intermediate features emerge. These results prove that conceptual attributes are untied and bound at distinct, predictable temporal phases, unlocking targeted acceleration strategies for large-scale generative architectures.

% \begin{figure}[h]
%     \centering
%     \input{images/figure_architecture}
%     \caption{Two-stage model architecture (VAE + U-Net)}
%     \label{fig:architecture}
% \end{figure}

% \begin{figure}[h]
%     \centering
%     \input{images/figure_sampling}
%     \caption{Sampling algorithm: Euler ODE + superposition decomposition}
%     \label{fig:sampling_algorithm}
% \end{figure}

%% file: images/fig_ood_mnist.tex
% fig_ood_mnist.tex — OOD generalization: 8 training pairs → 4 novel pairs
% Usage: \input{fig_ood_mnist} inside a figure environment
% Requires: fig_preamble.tex loaded in document preamble
%
% Designed for Springer LNCS textwidth (~12.2cm)

\begin{tikzpicture}[
  node distance=0.3cm,
  pairbox/.style={
    draw, rounded corners=3pt, minimum height=0.55cm,
    minimum width=1.35cm, align=center, font=\sffamily, thick,
    fill=white, inner sep=3pt
  },
  embnode/.style={
    draw, rounded corners=2pt, minimum height=0.38cm,
    minimum width=0.42cm, align=center,
    font=\sffamily\bfseries\scriptsize, thick, inner sep=1.5pt
  },
]

% ============================================================
% LEFT BLOCK: Training Pairs (4 rows × 2 columns)
% ============================================================
\begin{scope}[on background layer]
  \fill[vaelight, rounded corners=6pt]
    (-0.55, -4.15) rectangle (4.35, 0.95);
\end{scope}
\node[font=\scriptsize\bfseries\sffamily, text=vaecol]
  at (1.9, 0.68) {Observed Training Pairs};

% Row labels
\node[font=\tiny\sffamily, text=dig2col, anchor=east] at (0.4, 0.0)
  {$d_1\!=\!2$};
\node[font=\tiny\sffamily, text=dig3col, anchor=east] at (0.4, -1.1)
  {$d_1\!=\!3$};
\node[font=\tiny\sffamily, text=dig5col, anchor=east] at (0.4, -2.2)
  {$d_1\!=\!5$};
\node[font=\tiny\sffamily, text=dig8col, anchor=east] at (0.4, -3.3)
  {$d_1\!=\!8$};

% Row 1: 2x
\node[pairbox, draw=vaecol!60] (t23) at (1.1, 0.0)
  {\textcolor{dig2col}{\textbf{2}}\;\textcolor{dig3col}{\textbf{3}}};
\node[pairbox, draw=vaecol!60] (t25) at (2.9, 0.0)
  {\textcolor{dig2col}{\textbf{2}}\;\textcolor{dig5col}{\textbf{5}}};

% Row 2: 3x
\node[pairbox, draw=vaecol!60] (t35) at (1.1, -1.1)
  {\textcolor{dig3col}{\textbf{3}}\;\textcolor{dig5col}{\textbf{5}}};
\node[pairbox, draw=vaecol!60] (t38) at (2.9, -1.1)
  {\textcolor{dig3col}{\textbf{3}}\;\textcolor{dig8col}{\textbf{8}}};

% Row 3: 5x
\node[pairbox, draw=vaecol!60] (t58) at (1.1, -2.2)
  {\textcolor{dig5col}{\textbf{5}}\;\textcolor{dig8col}{\textbf{8}}};
\node[pairbox, draw=vaecol!60] (t52) at (2.9, -2.2)
  {\textcolor{dig5col}{\textbf{5}}\;\textcolor{dig2col}{\textbf{2}}};

% Row 4: 8x
\node[pairbox, draw=vaecol!60] (t82) at (1.1, -3.3)
  {\textcolor{dig8col}{\textbf{8}}\;\textcolor{dig2col}{\textbf{2}}};
\node[pairbox, draw=vaecol!60] (t83) at (2.9, -3.3)
  {\textcolor{dig8col}{\textbf{8}}\;\textcolor{dig3col}{\textbf{3}}};

% Thin horizontal separators between row groups
\draw[black!12, thin] (-0.35, -0.55) -- (4.15, -0.55);
\draw[black!12, thin] (-0.35, -1.65) -- (4.15, -1.65);
\draw[black!12, thin] (-0.35, -2.75) -- (4.15, -2.75);

% ============================================================
% CENTER: Disentangled Embeddings
% ============================================================
\begin{scope}[on background layer]
  \fill[disentlight, rounded corners=6pt]
    (5.0, -4.15) rectangle (7.9, 0.95);
\end{scope}
\node[font=\scriptsize\bfseries\sffamily, text=disent, align=center]
  at (6.45, 0.68) {Disentangled};
\node[font=\scriptsize\bfseries\sffamily, text=disent, align=center]
  at (6.45, 0.35) {Embeddings};

% Left-position embedding table
\node[blk, fill=disent!8, draw=disent!50, minimum width=2.2cm,
      minimum height=0.4cm, font=\tiny\sffamily]
  (embL) at (6.45, -0.15) {$\mathrm{Emb}_L$};

\node[embnode, fill=dig2col!12, draw=dig2col]
  (eL2) at (5.65, -0.7) {\textcolor{dig2col}{2}};
\node[embnode, fill=dig3col!12, draw=dig3col]
  (eL3) at (6.18, -0.7) {\textcolor{dig3col}{3}};
\node[embnode, fill=dig5col!12, draw=dig5col]
  (eL5) at (6.72, -0.7) {\textcolor{dig5col}{5}};
\node[embnode, fill=dig8col!12, draw=dig8col]
  (eL8) at (7.25, -0.7) {\textcolor{dig8col}{8}};

% Cartesian product
\node[font=\normalsize, text=disent!50] at (6.45, -1.25)
  {$\boldsymbol{\times}$};
\node[font=\tiny\sffamily, text=black!40, anchor=west]
  at (6.85, -1.25) {\textit{recombine}};

% Right-position embedding table
\node[blk, fill=disent!8, draw=disent!50, minimum width=2.2cm,
      minimum height=0.4cm, font=\tiny\sffamily]
  (embR) at (6.45, -1.75) {$\mathrm{Emb}_R$};

\node[embnode, fill=dig2col!12, draw=dig2col]
  (eR2) at (5.65, -2.3) {\textcolor{dig2col}{2}};
\node[embnode, fill=dig3col!12, draw=dig3col]
  (eR3) at (6.18, -2.3) {\textcolor{dig3col}{3}};
\node[embnode, fill=dig5col!12, draw=dig5col]
  (eR5) at (6.72, -2.3) {\textcolor{dig5col}{5}};
\node[embnode, fill=dig8col!12, draw=dig8col]
  (eR8) at (7.25, -2.3) {\textcolor{dig8col}{8}};

% Factored conditioning note
\node[font=\tiny\sffamily, text=disent, align=center]
  at (6.45, -3.15)
  {$c = \mathrm{Emb}_L(d_1) \,\|\, \mathrm{Emb}_R(d_2)$};
\node[font=\tiny\sffamily, text=black!45, align=center]
  at (6.45, -3.6)
  {each digit learned};
\node[font=\tiny\sffamily, text=black!45, align=center]
  at (6.45, -3.9)
  {independently per position};

% ============================================================
% RIGHT BLOCK: Novel OOD Pairs
% ============================================================
\begin{scope}[on background layer]
  \fill[oodlight, rounded corners=6pt]
    (8.55, -4.15) rectangle (11.15, 0.95);
\end{scope}
\node[font=\scriptsize\bfseries\sffamily, text=supercol!80!black]
  at (9.85, 0.68) {Novel OOD Pairs};

\node[pairbox, draw=supercol, fill=supercol!8, line width=1.2pt]
  (o28) at (9.85, 0.0)
  {\textcolor{dig2col}{\textbf{2}}\;\textcolor{dig8col}{\textbf{8}}};
\node[pairbox, draw=supercol, fill=supercol!8, line width=1.2pt]
  (o32) at (9.85, -1.1)
  {\textcolor{dig3col}{\textbf{3}}\;\textcolor{dig2col}{\textbf{2}}};
\node[pairbox, draw=supercol, fill=supercol!8, line width=1.2pt]
  (o53) at (9.85, -2.2)
  {\textcolor{dig5col}{\textbf{5}}\;\textcolor{dig3col}{\textbf{3}}};
\node[pairbox, draw=supercol, fill=supercol!8, line width=1.2pt]
  (o85) at (9.85, -3.3)
  {\textcolor{dig8col}{\textbf{8}}\;\textcolor{dig5col}{\textbf{5}}};

\node[font=\tiny\itshape\sffamily, text=supercol!65!black]
  at (9.85, -3.85) {unseen during training};

% ============================================================
% CONNECTING ARROWS
% ============================================================

% Training → Embeddings (thick sweep arrow)
\draw[-{Stealth[length=5pt, width=4pt]}, semithick, vaecol!55]
  (4.35, -1.65) -- (5.0, -1.65);

% Embeddings → OOD (thick sweep arrow)
\draw[-{Stealth[length=5pt, width=4pt]}, semithick, supercol!65]
  (7.9, -1.65) -- (8.55, -1.65);

% ============================================================
% SUBTLE PROVENANCE CURVES: show which training rows feed which OOD pairs
% (thin colored curves linking shared left-digit knowledge)
% ============================================================

% Row 2x trains left-2 → used in OOD "28"
\draw[dig2col!35, thin, densely dotted, bend left=18]
  (t25.east) to (o28.west);

% Row 3x trains left-3 → used in OOD "32"
\draw[dig3col!35, thin, densely dotted, bend left=12]
  (t38.east) to (o32.west);

% Row 5x trains left-5 → used in OOD "53"
\draw[dig5col!35, thin, densely dotted, bend left=12]
  (t52.east) to (o53.west);

% Row 8x trains left-8 → used in OOD "85"
\draw[dig8col!35, thin, densely dotted, bend left=18]
  (t83.east) to (o85.west);

\end{tikzpicture}

%% file: background.tex
\section{Background}

At the heart of our investigation lies a fundamental tension between how generative models learn distributions and how we expect them to compose concepts. Historically, compositionality was treated as a static architectural constraint or a latent space entanglement problem. However, recent paradigms in Energy-Based Models (EBMs) \cite{hinton2002training,shazeer2017outrageously,greff2019multi,du2021unsupervised} and continuous-time generative frameworks \cite{song2021denoising,liu2022compositional} suggest that composition is inherently probabilistic and dynamic.

The theoretical foundation for our exact compositional scoring originates from the product-of-experts formulation in EBMs. To combine independent concepts---such as a left digit $L$ and a right digit $R$---we assume their joint probability factorizes proportionally to their individual marginals over the unconditional prior: 

\begin{equation}
    p(x \mid L, R) \propto p(x) \frac{p(x \mid L)}{p(x)} \frac{p(x \mid R)}{p(x)}
\end{equation}

In the energy domain, where a distribution is defined by $p(x) \propto e^{-E(x)}$, this relationship elegantly collapses into a simple linear addition of their respective energy functions.

Composable Diffusion \cite{liu2022compositional} directly maps this static EBM logic onto the dynamic generative trajectory. Because the score network functionally parameterizes the gradient of the data distribution's log-density, the EBM product rule dictates the update step. For a generic vector field $v_\theta(z_t, t, c)$---which represents either the noise prediction $\epsilon_\theta$ in diffusion SDEs or the velocity field in Conditional Flow Matching ODEs---standard classifier-free guidance (CFG) extrapolates away from the unconditional prediction $\emptyset$ scaled by a guidance weight $\omega$:

\begin{equation}
    \tilde{v}_\theta(z_t, t, c) = v_\theta(z_t, t, \emptyset) + \omega \left( v_\theta(z_t, t, c) - v_\theta(z_t, t, \emptyset) \right)
\end{equation}

When a model is tasked with generating a multi-concept image, the conventional strategy relies on a \textit{joint score}. The network is conditioned on the concatenated concepts simultaneously, relying entirely on the assumption that the model has learned a perfectly disentangled representation during training:

\begin{equation}
   \tilde{v}_{\text{joint}} = v_\theta(z_t, t, \emptyset) + \omega \big( v_\theta(z_t, t, L, R) - v_\theta(z_t, t, \emptyset) \big) 
\end{equation}

While computationally efficient (requiring only two forward passes per step), joint scoring frequently collapses when faced with out-of-distribution combinations. The mathematically rigorous alternative is the \textit{decomposed score}. By explicitly enforcing the log-gradient factorization derived from Composable Diffusion, we query the network independently for the left and right concepts before assembling their guidance vectors:

\begin{equation}
    \tilde{v}_{\text{decomp}} = v_\theta(z_t, t, \emptyset) + \omega \big( v_\theta(z_t, t, L, \emptyset) + v_\theta(z_t, t, \emptyset, R) - 2 v_\theta(z_t, t, \emptyset) \big)
\end{equation}

This explicit factorization is highly reliable for OOD generalization, yet the computational penalty is severe. Evaluating the decomposed score demands three distinct forward passes per integration step. Reflecting on this overhead, we hypothesized that this rigid, continuous mathematical enforcement is largely redundant. It is highly improbable that the network actively binds independent concepts uniformly across the entire generation trajectory; rather, there must exist a specific, critical temporal phase where these independent vector fields actually interfere, interact, and resolve.

%% file: bifurcation_window.tex
\section{Semantic Bifurcation Window and Surgical Guidance}

We questioned whether the model actively struggles with compositional binding throughout the entire generation process. If the network only negotiates conceptual conflicts during a specific phase, enforcing a rigorous and computationally expensive decomposed score at every integration step is inherently wasteful. To investigate this, we quantified the functional discrepancy---which we term \textit{interference}---between the joint and decomposed vector fields. 

By tracking the Mean Squared Error (MSE) and Cosine Similarity $\mathcal{S}_{\text{cos}}$ $(\tilde{v}_{\text{joint}}, \tilde{v}_{\text{decomp}})$ across the normalized sampling trajectory, a striking dynamic pattern emerged.

\begin{figure}[!htbp]
\includegraphics[width=\textwidth]{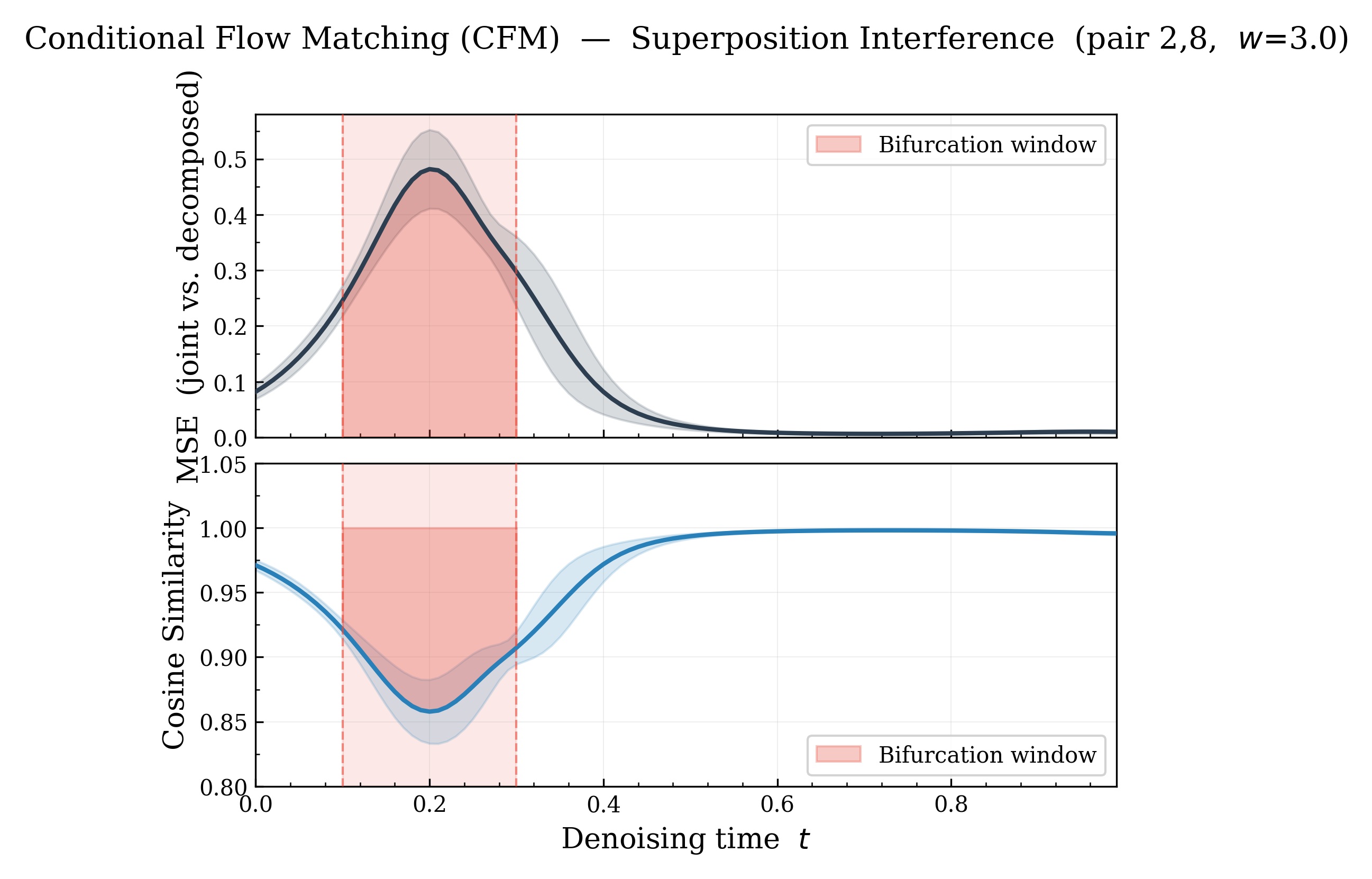}
\includegraphics[width=\textwidth]{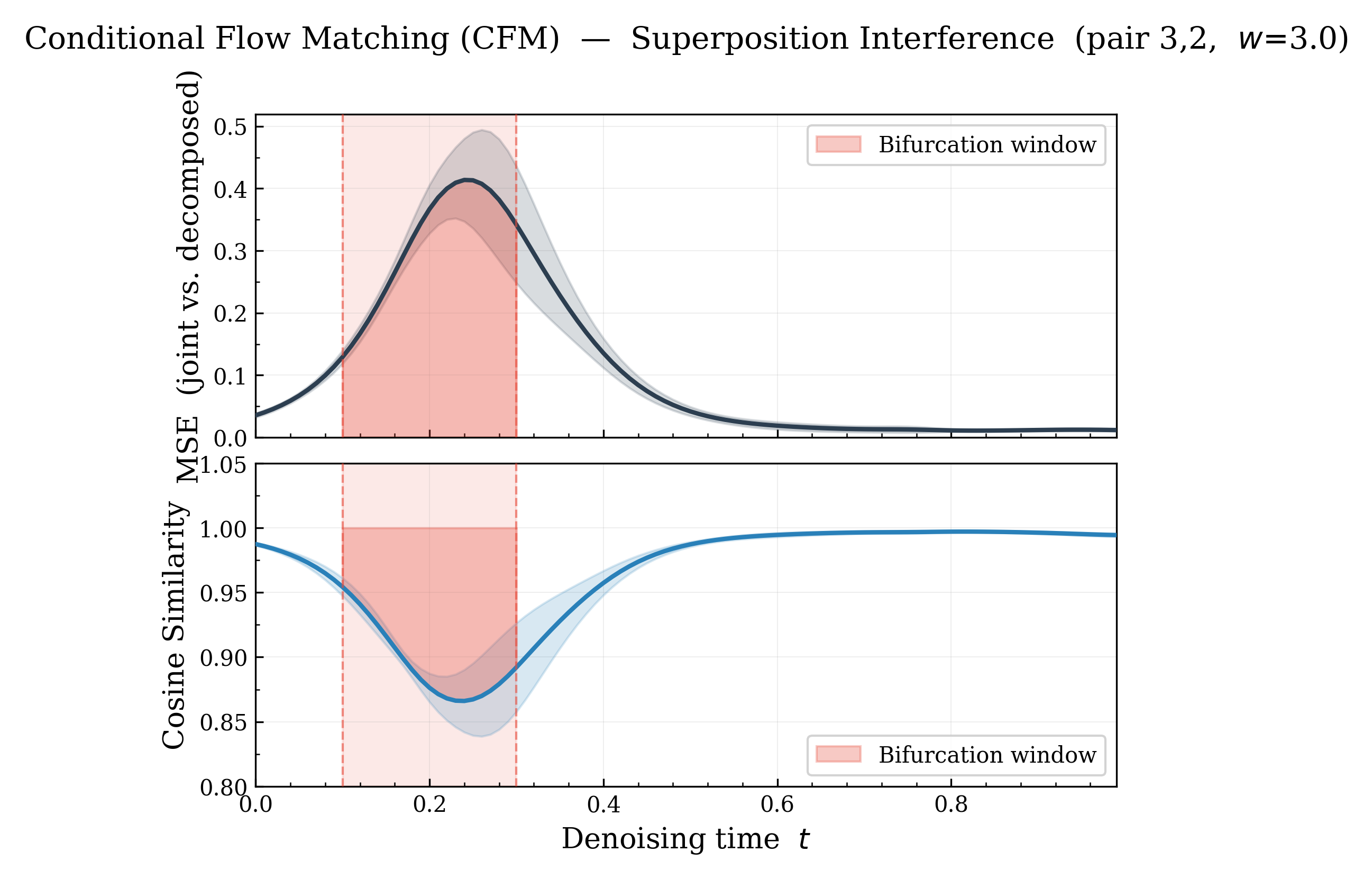}
\caption{Divergence between joint and decomposed scoring vector fields.}
\end{figure}

The two scoring strategies dictate nearly identical trajectories at the extremes of the generation process. Early on, the latent state is dominated by structureless noise; late in the process, the topology is already rigidly established. However, the vector fields diverge drastically during a highly localized intermediate phase. We formalize this critical region of instability as the \textit{semantic bifurcation window}, denoted by the interval $[t_{\text{start}}, t_{\text{end}}]$. It is exclusively within this temporal pocket that the network is forced to untangle and resolve the conflicting spatial gradients of the independent digit embeddings.

\begin{figure}[!htbp]
\includegraphics[width=\textwidth]{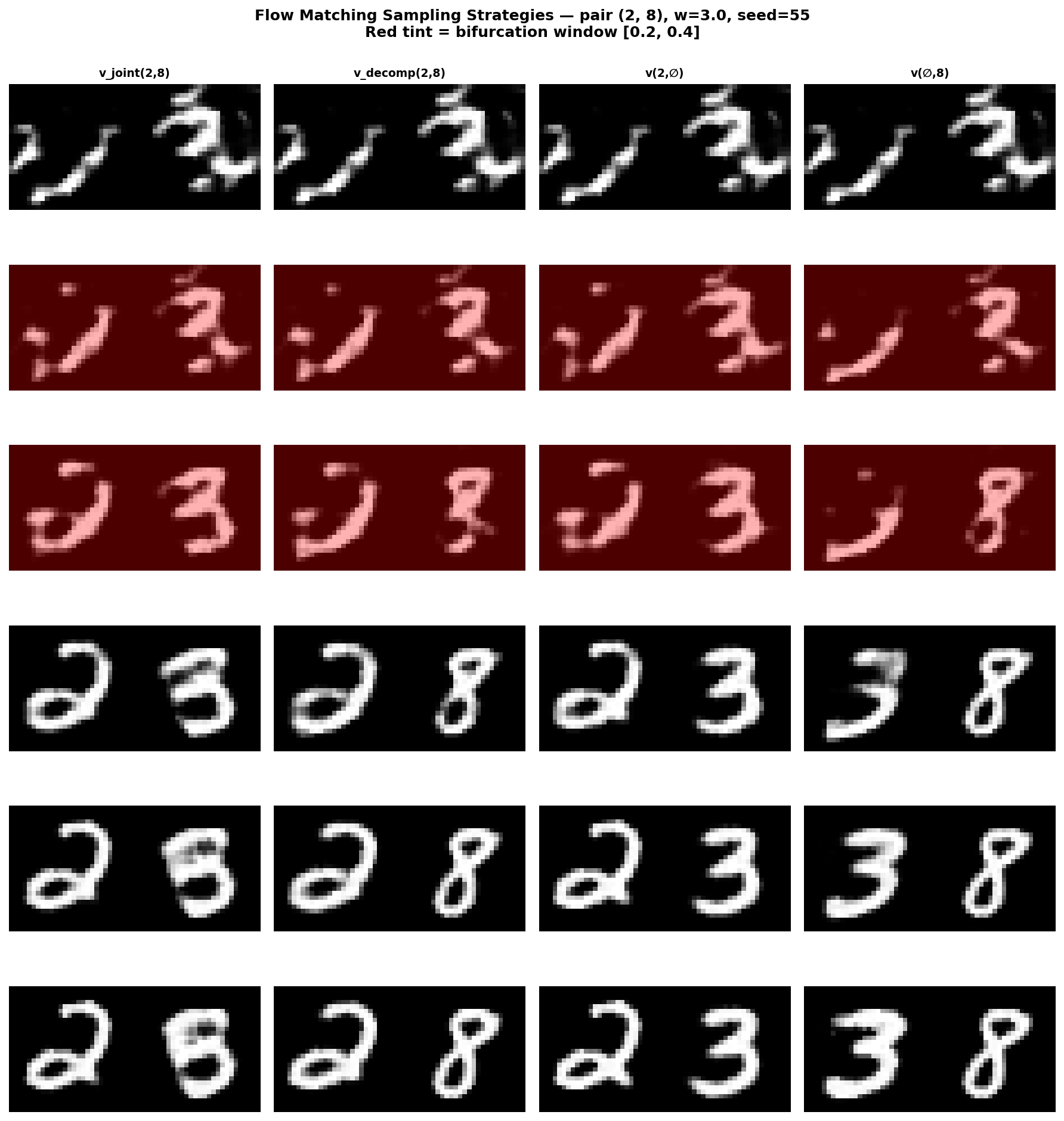}
\caption{Visualizing attractor collapse during the semantic bifurcation window.}
\end{figure}

\begin{table}[!ht]
\centering
\caption{\textbf{Out-of-distribution compositional generalization errors.} Values represent the number of misclassified samples out of 300 novel held-out pairs (lower is better).}
\begin{tabular}{lccccccccc}
\toprule
Method & \textbf{(2,8)} & \textbf{(3,2)} & \textbf{(5,3)} & \textbf{(8,5)} & \textbf{(2,2)} & \textbf{(3,3)} & \textbf{(5,5)} & \textbf{(8,8)} & Total \\
\midrule
CFDG (joint) & 74 & 47 & 45 & 57 & 32 & 12 & 50 & 6 & 323 \\
CFDG (decomposed) & 27 & 28 & 29 & 30 & 15 & 3 & 37 & 2 & \textbf{171} \\
CFM (joint) & 19 & 23 & 28 & 21 & 14 & 14 & 18 & 21 & 158 \\
CFM (decomposed) & 18 & 21 & 9 & 14 & 8 & 2 & 15 & 12 & \textbf{99} \\
\bottomrule
\end{tabular}
\vspace{0.2cm}
\label{tab:ood-errors}
\end{table}

This observation fundamentally alters how we can approach compositional sampling. If the standard joint score naturally approximates the decomposed score outside of this active window, we can bypass the three-pass mathematical factorization entirely for the majority of the trajectory. This led us to formulate \textbf{surgical guidance}, a hybrid sampling strategy that dynamically switches the guidance objective based on the current timestep:

\begin{equation}
    \tilde{v}_{\text{surgical}}(z_t, t) = \begin{cases} \tilde{v}_{\text{decomp}}(z_t, t) & \text{if } t \in [t_{\text{start}}, t_{\text{end}}] \\ \tilde{v}_{\text{joint}}(z_t, t) & \text{otherwise} \end{cases}
\end{equation}

By restricting the exact decomposed evaluation strictly to the semantic bifurcation window, surgical guidance effectively insulates the generative process against out-of-distribution collapse. This targeted intervention guarantees the structural integrity of novel concept combinations while preserving the computational efficiency of standard classifier-free guidance everywhere else.

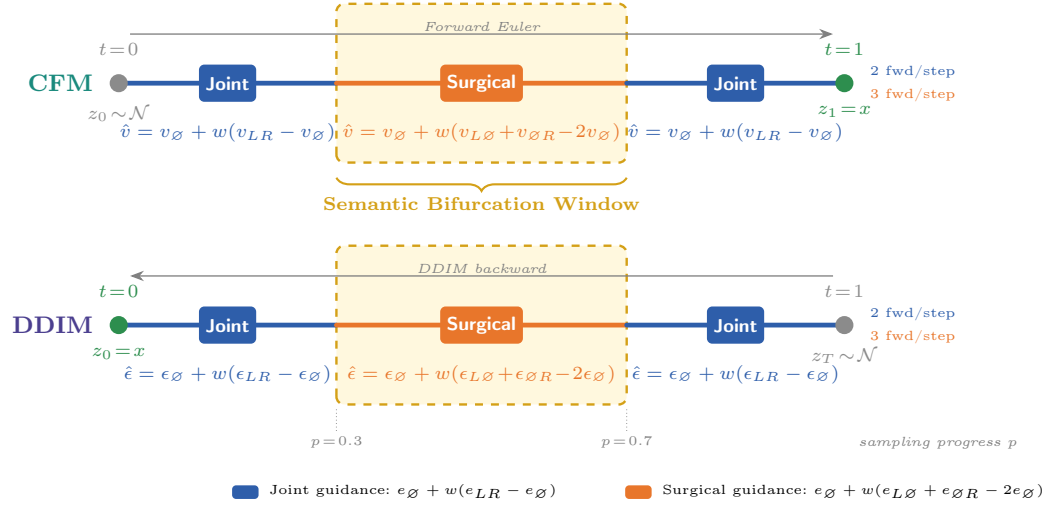
\begin{figure}[!htbp]
    \centering
    \input{images/figure_surgical_guidance}
    \caption{%
      Surgical guidance mechanism across CFM and DDIM frameworks.
    }
    \label{fig:surgical_guidance}
\end{figure}

\begin{algorithm}[!htbp]
\caption{Surgical Guidance for Conditional Flow Matching (CFM)}
\label{alg:surgical_guidance_cfm}
\begin{algorithmic}[1]
\Require Initial noise $z_0 \sim \mathcal{N}(0, I)$, concepts $(c_L, c_R)$, bifurcation window $[\tau_{\text{start}}, \tau_{\text{end}}]$, integration steps $N$, guidance scale $\omega$
\Ensure Generated image $x^*$
\State $\Delta t \gets 1 / N$
\State $z \gets z_0$
\For{$i = 0$ \textbf{to} $N-1$}
    \State $t \gets i \cdot \Delta t$
    \State $\text{prog} \gets i / N$
    \State $v_{\text{uncond}} \gets v_\theta(z, t, \emptyset, \emptyset)$
    \If{$\tau_{\text{start}} \le \text{prog} \le \tau_{\text{end}}$} \Comment{\textit{Semantic Bifurcation Window (3 NFEs)}}
        \State $v_{\text{left}} \gets v_\theta(z, t, c_L, \emptyset)$
        \State $v_{\text{right}} \gets v_\theta(z, t, \emptyset, c_R)$
        \State $\tilde{v}_t \gets v_{\text{uncond}} + \omega \cdot (v_{\text{left}} + v_{\text{right}} - 2 \cdot v_{\text{uncond}})$
    \Else \Comment{\textit{Standard Joint CFG (2 NFEs)}}
        \State $v_{\text{cond}} \gets v_\theta(z, t, c_L, c_R)$
        \State $\tilde{v}_t \gets v_{\text{uncond}} + \omega \cdot (v_{\text{cond}} - v_{\text{uncond}})$
    \EndIf
    \State $z \gets z + \tilde{v}_t \cdot \Delta t$ \Comment{\textit{Forward Euler ODE step}}
\EndFor
\State $x^* \gets \text{Decoder}(z)$
\State \Return $x^*$
\end{algorithmic}
\end{algorithm}

\begin{algorithm}[!ht]
\caption{Surgical Guidance for Diffusion Models (DDIM)}
\label{alg:surgical_guidance_diffusion}
\begin{algorithmic}[1]
\Require Initial noise $z_N \sim \mathcal{N}(0, I)$, concepts $(c_L, c_R)$, bifurcation window $[\tau_{\text{start}}, \tau_{\text{end}}]$, integration steps $N$, guidance scale $\omega$
\Ensure Generated image $x^*$
\State $\Delta t \gets 1 / N$
\State $z \gets z_N$
\For{$i = 0$ \textbf{to} $N-1$}
    \State $t \gets 0.999 - i \cdot \Delta t$
    \State $t_{\text{next}} \gets t - \Delta t$
    \State $\text{prog} \gets i / N$
    \State $\epsilon_{\text{uncond}} \gets \epsilon_\theta(z, t, \emptyset, \emptyset)$
    \If{$\tau_{\text{start}} \le \text{prog} \le \tau_{\text{end}}$} \Comment{\textit{Semantic Bifurcation Window (3 NFEs)}}
        \State $\epsilon_{\text{left}} \gets \epsilon_\theta(z, t, c_L, \emptyset)$
        \State $\epsilon_{\text{right}} \gets \epsilon_\theta(z, t, \emptyset, c_R)$
        \State $\tilde{\epsilon}_t \gets \epsilon_{\text{uncond}} + \omega \cdot (\epsilon_{\text{left}} + \epsilon_{\text{right}} - 2 \cdot \epsilon_{\text{uncond}})$
    \Else \Comment{\textit{Standard Joint CFG (2 NFEs)}}
        \State $\epsilon_{\text{cond}} \gets \epsilon_\theta(z, t, c_L, c_R)$
        \State $\tilde{\epsilon}_t \gets \epsilon_{\text{uncond}} + \omega \cdot (\epsilon_{\text{cond}} - \epsilon_{\text{uncond}})$
    \EndIf
    \State $z_0^{\text{pred}} \gets \frac{z - \sqrt{1 - \bar{\alpha}_t} \cdot \tilde{\epsilon}_t}{\sqrt{\bar{\alpha}_t}}$
    \State $z \gets \sqrt{\bar{\alpha}_{t_{\text{next}}}} \cdot z_0^{\text{pred}} + \sqrt{1 - \bar{\alpha}_{t_{\text{next}}}} \cdot \tilde{\epsilon}_t$ \Comment{\textit{DDIM step}}
\EndFor
\State $x^* \gets \text{Decoder}(z)$
\State \Return $x^*$
\end{algorithmic}
\end{algorithm}

%% file: images/figure_surgical_guidance.tex
% Surgical Guidance (Hybrid Sampling) — Illustration
% Fits Springer LNCS text width (~122mm / 12.2cm)
% Include surgical_guidance_preamble.tex in your document preamble

% \begin{figure}[t]
% \centering
% \resizebox{\textwidth}{!}{%
\begin{tikzpicture}[
    >=Stealth,
    font=\footnotesize,
    every node/.style={inner sep=1pt},
    timeline/.style={line width=1.8pt, line cap=round},
    phase label/.style={font=\scriptsize\bfseries, text=white,
                        inner sep=2.5pt, rounded corners=2pt,
                        minimum height=14pt},
    formula/.style={font=\scriptsize, text=black, align=center},
    heading/.style={font=\small\bfseries, anchor=east},
    endpoint/.style={circle, minimum size=6pt, inner sep=0pt,
                     line width=0.8pt},
    brace label/.style={font=\scriptsize\bfseries, text=bifBorder},
    fwd label/.style={font=\tiny, text=noiseGray, align=center},
  ]

  % === Dimensions ===
  \def\tlLen{9.6}    % timeline length (cm)
  \def\tlY{0}        % y-offset base
  \def\rowSep{3.2}   % vertical separation between two model rows
  \def\bifS{0.30}    % bifurcation start (fraction)
  \def\bifE{0.70}    % bifurcation end (fraction)

  % === Computed positions ===
  \pgfmathsetmacro{\xStart}{0}
  \pgfmathsetmacro{\xEnd}{\tlLen}
  \pgfmathsetmacro{\xBifS}{\tlLen * \bifS}
  \pgfmathsetmacro{\xBifE}{\tlLen * \bifE}

  % =====================================================================
  %   CFM  (Conditional Flow Matching) — top row
  % =====================================================================
  \begin{scope}[yshift=\rowSep cm]

    % --- Bifurcation window background ---
    \fill[bifYellow, rounded corners=3pt, opacity=0.55]
      (\xBifS, -1.05) rectangle (\xBifE, 1.05);
    \draw[bifBorder, line width=0.9pt, dashed, rounded corners=3pt]
      (\xBifS, -1.05) rectangle (\xBifE, 1.05);

    % --- Row heading ---
    \node[heading, text=cfmTeal] at (-0.3, 0) {CFM};

    % --- Timeline segments ---
    % Joint pre-window
    \draw[timeline, jointBlue]
      (\xStart, 0) -- (\xBifS, 0);
    % Decomposed (surgical) window
    \draw[timeline, decompOrange]
      (\xBifS, 0) -- (\xBifE, 0);
    % Joint post-window
    \draw[timeline, jointBlue]
      (\xBifE, 0) -- (\xEnd, 0);

    % --- Endpoints ---
    \node[endpoint, fill=noiseGray, draw=noiseGray]  (cfm0) at (\xStart, 0) {};
    \node[endpoint, fill=dataGreen, draw=dataGreen]  (cfm1) at (\xEnd, 0) {};

    % Endpoint labels
    \node[below=4pt of cfm0, font=\scriptsize\bfseries, text=noiseGray] {$z_0 \!\sim\! \mathcal{N}$};
    \node[below=4pt of cfm1, font=\scriptsize\bfseries, text=dataGreen] {$z_1 \!=\! x$};

    % --- Time axis labels ---
    \node[above=6pt of cfm0, font=\scriptsize, text=noiseGray] {$t\!=\!0$};
    \node[above=6pt of cfm1, font=\scriptsize, text=dataGreen] {$t\!=\!1$};

    % --- Direction arrow ---
    \draw[->, gray, line width=0.5pt]
      (\xStart+0.15, 0.65) -- (\xEnd-0.15, 0.65)
      node[midway, above, font=\tiny\itshape, text=gray] {Forward Euler};

    % --- Phase labels (on timeline) ---
    \node[phase label, fill=jointBlue]
      at ({(\xStart+\xBifS)/2}, 0) {\textsf{Joint}};
    \node[phase label, fill=decompOrange, minimum width=20pt]
      at ({(\xBifS+\xBifE)/2}, 0) {\textsf{Surgical}};
    \node[phase label, fill=jointBlue]
      at ({(\xBifE+\xEnd)/2}, 0) {\textsf{Joint}};

    % --- Formulas below timeline ---
    \node[formula, text=jointBlue] at ({(\xStart+\xBifS)/2}, -0.65)
      {$\hat{v} = v_\varnothing + w(v_{LR} - v_\varnothing)$};
    \node[formula, text=decompOrange] at ({(\xBifS+\xBifE)/2}, -0.65)
      {$\hat{v} = v_\varnothing + w(v_{L\varnothing} \!+\! v_{\varnothing R} \!-\! 2v_\varnothing)$};
    \node[formula, text=jointBlue] at ({(\xBifE+\xEnd)/2}, -0.65)
      {$\hat{v} = v_\varnothing + w(v_{LR} - v_\varnothing)$};

  \end{scope}

  % =====================================================================
  %   DDIM  (Denoising Diffusion) — bottom row
  % =====================================================================
  \begin{scope}[yshift=0cm]

    % --- Bifurcation window background ---
    \fill[bifYellow, rounded corners=3pt, opacity=0.55]
      (\xBifS, -1.05) rectangle (\xBifE, 1.05);
    \draw[bifBorder, line width=0.9pt, dashed, rounded corners=3pt]
      (\xBifS, -1.05) rectangle (\xBifE, 1.05);

    % --- Row heading ---
    \node[heading, text=ddimIndigo] at (-0.3, 0) {DDIM};

    % --- Timeline segments ---
    \draw[timeline, jointBlue]
      (\xStart, 0) -- (\xBifS, 0);
    \draw[timeline, decompOrange]
      (\xBifS, 0) -- (\xBifE, 0);
    \draw[timeline, jointBlue]
      (\xBifE, 0) -- (\xEnd, 0);

    % --- Endpoints ---
    \node[endpoint, fill=dataGreen, draw=dataGreen]  (ddim0) at (\xStart, 0) {};
    \node[endpoint, fill=noiseGray, draw=noiseGray]  (ddim1) at (\xEnd, 0) {};

    % Endpoint labels
    \node[below=4pt of ddim0, font=\scriptsize\bfseries, text=dataGreen] {$z_0 \!=\! x$};
    \node[below=4pt of ddim1, font=\scriptsize\bfseries, text=noiseGray] {$z_T \!\sim\! \mathcal{N}$};

    % --- Time axis labels ---
    \node[above=6pt of ddim0, font=\scriptsize, text=dataGreen] {$t\!=\!0$};
    \node[above=6pt of ddim1, font=\scriptsize, text=noiseGray] {$t\!=\!1$};

    % --- Direction arrow (DDIM samples backward) ---
    \draw[<-, gray, line width=0.5pt]
      (\xStart+0.15, 0.65) -- (\xEnd-0.15, 0.65)
      node[midway, above, font=\tiny\itshape, text=gray] {DDIM backward};

    % --- Phase labels (on timeline) — note: reversed order for DDIM ---
    \node[phase label, fill=jointBlue]
      at ({(\xStart+\xBifS)/2}, 0) {\textsf{Joint}};
    \node[phase label, fill=decompOrange, minimum width=20pt]
      at ({(\xBifS+\xBifE)/2}, 0) {\textsf{Surgical}};
    \node[phase label, fill=jointBlue]
      at ({(\xBifE+\xEnd)/2}, 0) {\textsf{Joint}};

    % --- Formulas below timeline ---
    \node[formula, text=jointBlue] at ({(\xStart+\xBifS)/2}, -0.65)
      {$\hat{\epsilon} = \epsilon_\varnothing + w(\epsilon_{LR} - \epsilon_\varnothing)$};
    \node[formula, text=decompOrange] at ({(\xBifS+\xBifE)/2}, -0.65)
      {$\hat{\epsilon} = \epsilon_\varnothing + w(\epsilon_{L\varnothing} \!+\! \epsilon_{\varnothing R} \!-\! 2\epsilon_\varnothing)$};
    \node[formula, text=jointBlue] at ({(\xBifE+\xEnd)/2}, -0.65)
      {$\hat{\epsilon} = \epsilon_\varnothing + w(\epsilon_{LR} - \epsilon_\varnothing)$};

  \end{scope}

  % =====================================================================
  %   Shared annotations
  % =====================================================================

  % --- Bifurcation window brace (between the two rows) ---
  \draw[decorate, decoration={brace, amplitude=5pt, mirror}, bifBorder, line width=0.8pt]
    (\xBifS, {\rowSep - 1.25}) -- (\xBifE, {\rowSep - 1.25})
    node[midway, below=6pt, brace label] {Semantic Bifurcation Window};

  % --- Progress axis annotation ---
  \node[font=\tiny, text=gray] at (\xBifS, -1.55) {$p\!=\!0.3$};
  \node[font=\tiny, text=gray] at (\xBifE, -1.55) {$p\!=\!0.7$};
  \draw[gray, line width=0.3pt, densely dotted] (\xBifS, -1.4) -- (\xBifS, -1.1);
  \draw[gray, line width=0.3pt, densely dotted] (\xBifE, -1.4) -- (\xBifE, -1.1);

  % --- Progress label ---
  \node[font=\tiny\itshape, text=gray, anchor=west] at (\xEnd+0.15, -1.55)
    {sampling progress $p$};

  % --- Forward pass counts (right side) ---
  \node[fwd label, anchor=west, text=jointBlue] at (\xEnd+0.3, {\rowSep + 0.15})
    {2 fwd/step};
  \node[fwd label, anchor=west, text=decompOrange] at (\xEnd+0.3, {\rowSep - 0.15})
    {3 fwd/step};

  \node[fwd label, anchor=west, text=jointBlue] at (\xEnd+0.3, 0.15)
    {2 fwd/step};
  \node[fwd label, anchor=west, text=decompOrange] at (\xEnd+0.3, -0.15)
    {3 fwd/step};

  % --- Legend ---
  \begin{scope}[yshift=-2.3cm, xshift=1.5cm]
    \fill[jointBlue, rounded corners=1.5pt] (0, 0) rectangle (0.35, 0.18);
    \node[font=\tiny, anchor=west] at (0.45, 0.09)
      {Joint guidance: $e_\varnothing + w(e_{LR} - e_\varnothing)$};
    \fill[decompOrange, rounded corners=1.5pt] (5.2, 0) rectangle (5.55, 0.18);
    \node[font=\tiny, anchor=west] at (5.65, 0.09)
      {Surgical guidance: $e_\varnothing + w(e_{L\varnothing} + e_{\varnothing R} - 2e_\varnothing)$};
  \end{scope}

\end{tikzpicture}%
% } end resizebox
% \label{fig:surgical-guidance}
% \end{figure}

%% file: experiments.tex
\section{Experiments}

\subsection{Experimental Setup and Evaluation Metrics}

To empirically validate surgical guidance and locate the exact semantic bifurcation windows, we evaluate out-of-distribution generation on our controlled bi-digit MNIST testbed. We systematically sweep a sliding intervention window of fixed lengths ($l = 0.1$ and $l = 0.15$) across the normalized integration timeline. Within each designated window, we apply the exact three-pass decomposed score; outside of it, we revert to the standard two-pass joint score.

To rigorously quantify compositional failures, we employ a disentangled classifier that independently predicts the left and right digits of a generated image. A sample is strictly counted as a compositional error (Table \ref{tab:ood-errors}) if either spatial prediction diverges from the conditioning label.

Furthermore, to measure generative fidelity, we compute the Fréchet Inception Distance (FID) \cite{heusel2017gans} for surgical sampling on the out-of-distribution pairs (Table \ref{tab:fid-hybrid}). Features for the FID calculation are extracted from the penultimate layer of our disentangled digit classifier, yielding a 128-dimensional representation tailored to the spatial halves. Lower FID values indicate stronger distributional alignment with the ground-truth training set.

\subsection{Empirical Results and Topological Divergence}

Our evaluation measures the absolute improvement in OOD pairwise accuracy over the joint CFG baseline across 50 random seeds. The empirical results expose a stark topological divide between generative frameworks.

For diffusion models (SDEs), the joint baseline achieves an $87.2\%$ pairwise accuracy. As illustrated in Figures 5 and 6, the optimal semantic bifurcation window occurs aggressively early in the generation process. Applying surgical guidance strictly during the initial $10\%$ to $15\%$ of the trajectory (progress $\in [0.00, 0.15]$) yields a massive $+5.3\%$ absolute improvement. This indicates that the SDE framework forces compositional binding almost immediately out of pure noise, locking in structural geometry well before any high-frequency details are synthesized.

Conversely, Conditional Flow Matching (CFM) dictates a fundamentally different dynamic. CFM inherently possesses a stronger joint baseline of $94.2\%$. More importantly, its bifurcation window is temporally delayed. Figures 7 and 8 demonstrate that the optimal intervention zone peaks between progress bounds $[0.10, 0.25]$, delivering up to a $+1.1\%$ improvement. The ODE trajectory effectively defers structural composition. Rather than resolving conceptual conflicts at peak noise, flow matching waits until the latent state has partially smoothed into intermediate features before untangling the left and right positional embeddings.

Our quantitative FID analysis (Table \ref{tab:fid-hybrid}) reinforces these observations. CFDG surgical sampling attains a lower mean FID than CFM across all held-out compositions ($15.4$ vs. $16.3$), corroborating the error-count analysis. Interestingly, we observe an architectural quirk: cross-digit pairs consistently yield lower FIDs than repeated-digit compositions (e.g., $(k,k)$ pairs). This suggests that our localized surgical intervention is inherently more effective at disentangling semantically distinct digits than isolating identical, overlapping concepts.

\begin{table}[H]
\centering
\caption{\textbf{Fréchet Inception Distance (FID) for surgical sampling on out-of-distribution pairs.} Lower values indicate stronger distributional alignment with the training set.}
\begin{tabular}{lccccccccc}
\toprule
Method & \textbf{(2,8)} & \textbf{(3,2)} & \textbf{(5,3)} & \textbf{(8,5)} & \textbf{(2,2)} & \textbf{(3,3)} & \textbf{(5,5)} & \textbf{(8,8)} & Mean \\
\midrule
CFDG (surgical) & 16.9 & 14.5 & 13.5 & 13.9 & 10.5 & 18.7 & 11.8 & 23.5 & \textbf{15.4} \\
CFM (surgical) & 16.9 & 15.7 & 15.5 & 16.4 & 12.7 & 19.0 & 14.3 & 19.6 & \textbf{16.3} \\
\bottomrule
\end{tabular}
\vspace{0.2cm}
\label{tab:fid-hybrid}
\end{table}

\begin{figure}[h]
    \centering
    \begin{minipage}[t]{0.48\textwidth}
        \centering
        \includegraphics[width=\textwidth]{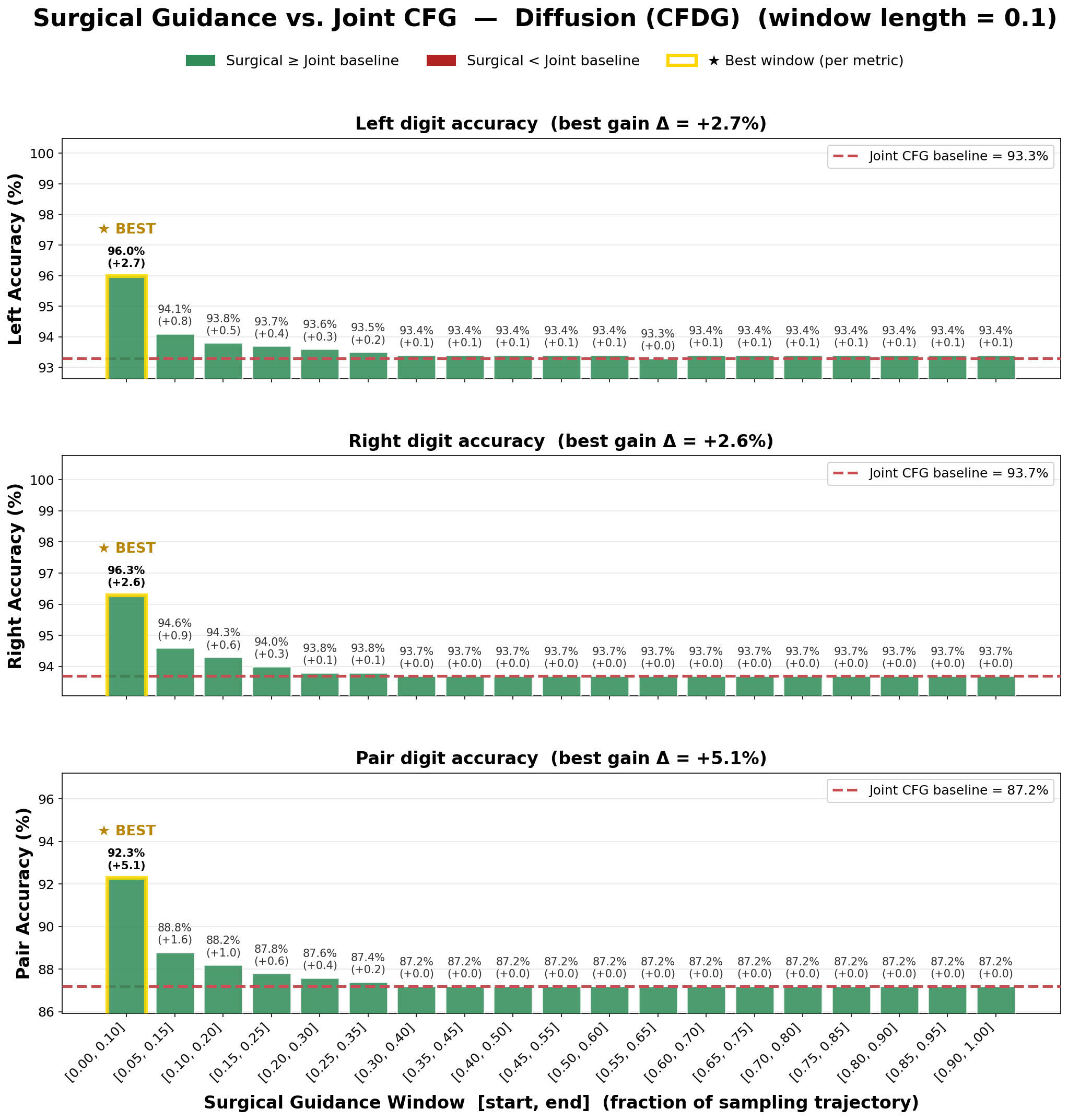}
        \caption{Diffusion bifurcation sweep ($l=0.1$). Optimal intervention occurs immediately at peak noise, yielding a $+5.1\%$ improvement.}
        \label{fig:bif_diff_01}
    \end{minipage}\hfill
    \begin{minipage}[t]{0.48\textwidth}
        \centering
        \includegraphics[width=\textwidth]{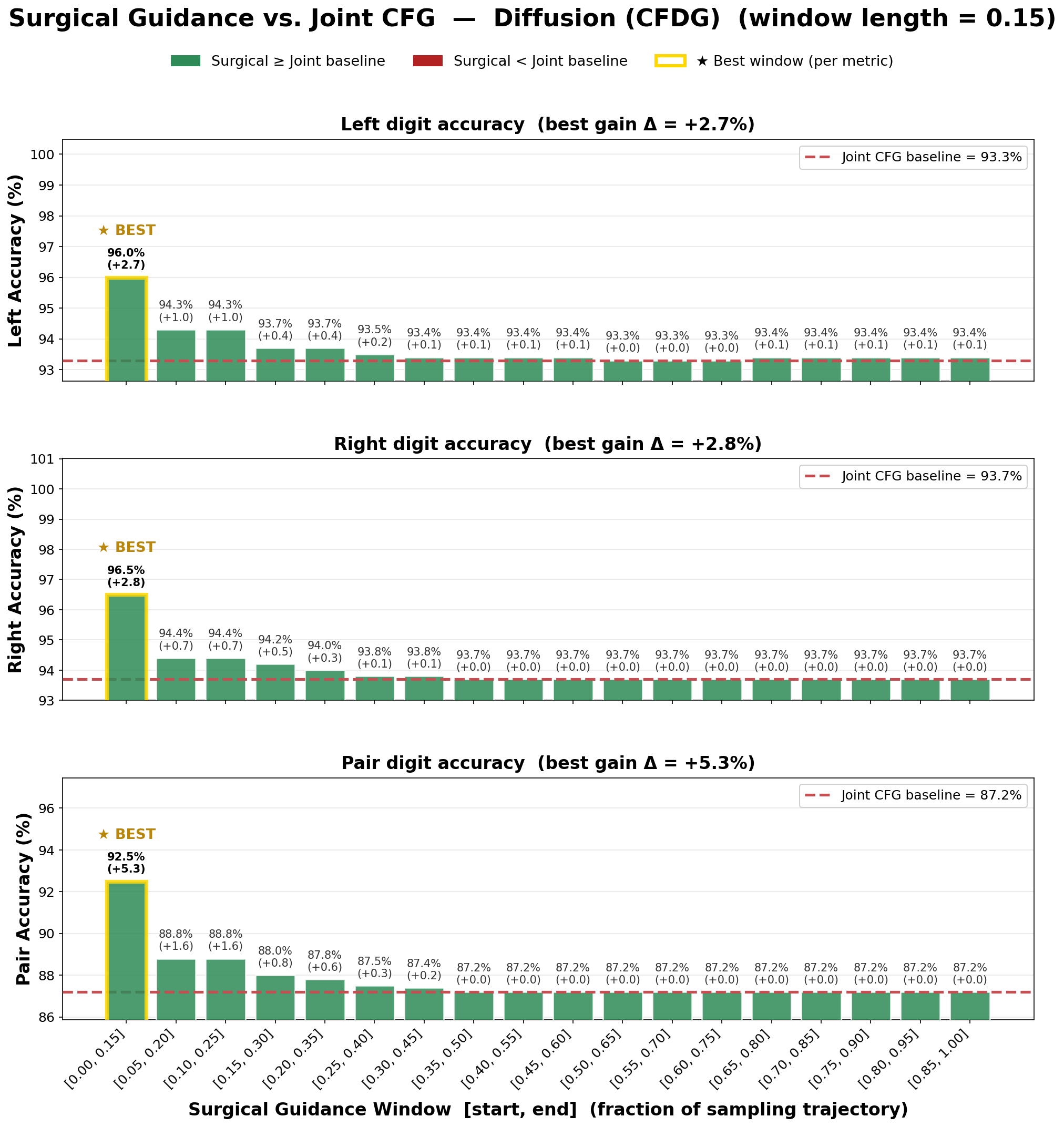}
        \caption{Diffusion bifurcation sweep ($l=0.15$). Expanding the window slightly captures a $+5.3\%$ gain, confirming early topological resolution.}
        \label{fig:bif_diff_015}
    \end{minipage}
\end{figure}

\begin{figure}[h]
    \centering
    \begin{minipage}[t]{0.48\textwidth}
        \centering
        \includegraphics[width=\textwidth]{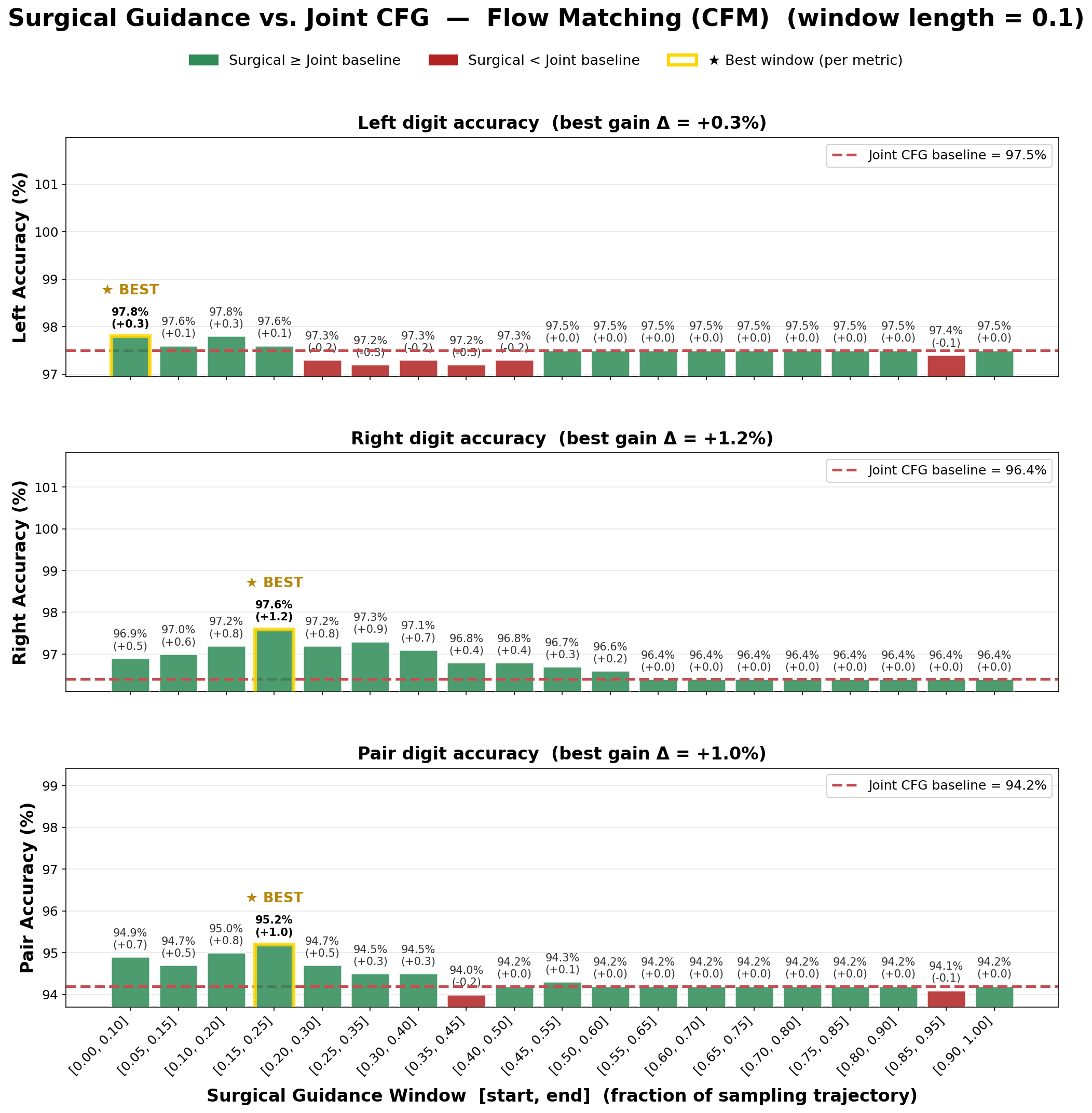}
        \caption{CFM bifurcation sweep ($l=0.1$). The semantic window is delayed, peaking at $[0.15, 0.25]$ to resolve compositional conflicts.}
        \label{fig:bif_flow_01}
    \end{minipage}\hfill
    \begin{minipage}[t]{0.48\textwidth}
        \centering
        \includegraphics[width=\textwidth]{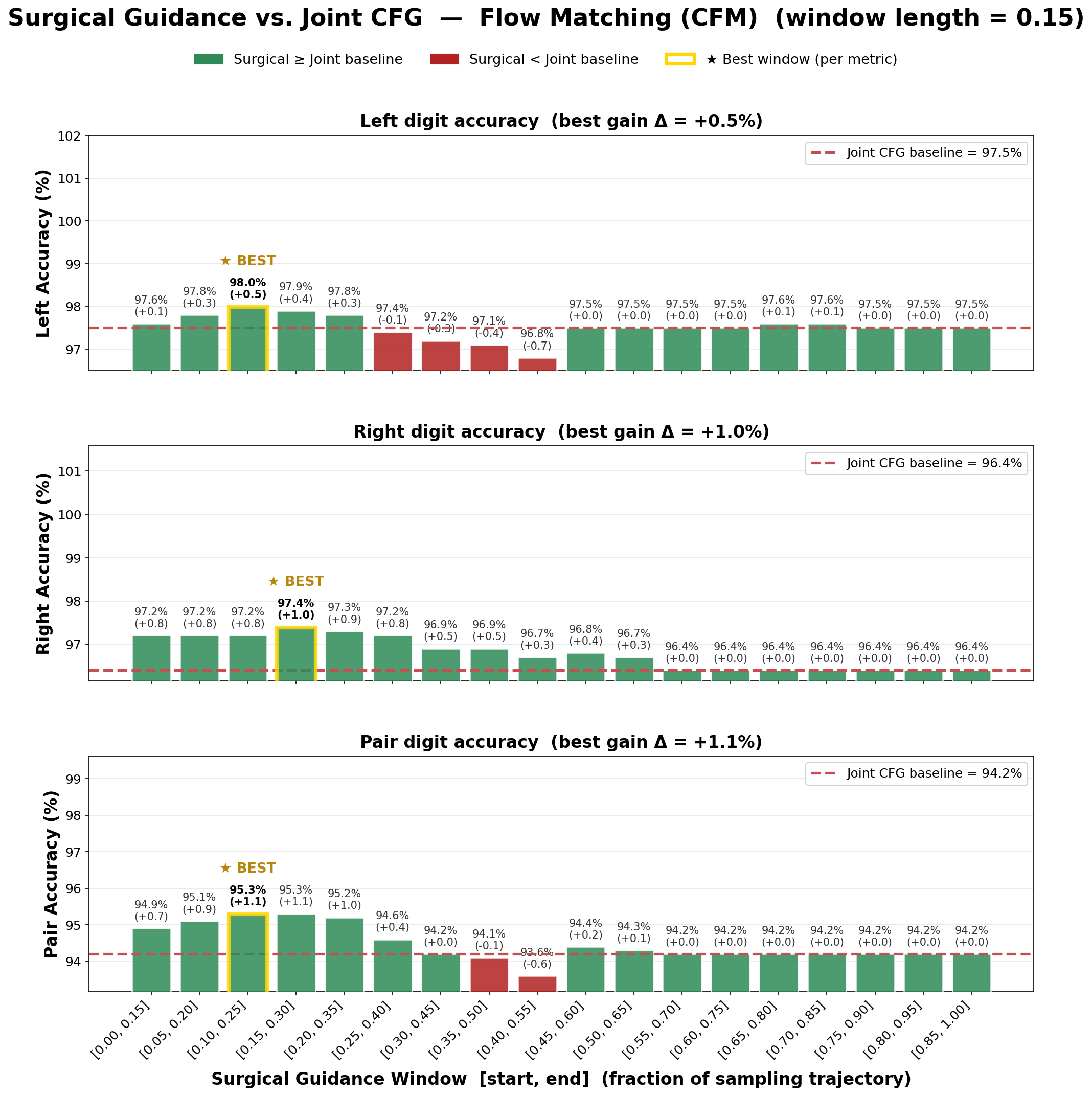}
        \caption{CFM bifurcation sweep ($l=0.15$). Optimal structural binding confirms the delayed ODE resolution phase.}
        \label{fig:bif_flow_015}
    \end{minipage}
\end{figure}

%% file: conclusion.tex
\section{Conclusion and Future Potential}

The identification of the semantic bifurcation window shifts the paradigm of compositional generation from a spatial representation problem to a dynamical systems problem. By isolating the exact temporal phase where generative models resolve conceptual interference, surgical guidance demonstrates that mathematically rigorous, multi-pass scoring is only necessary for a brief fraction of the sampling trajectory. This targeted intervention successfully preserves the out-of-distribution fidelity of full decomposed scoring while bypassing redundant neural evaluations elsewhere. 

Looking forward, the implications of temporal structural binding extend far beyond controlled testbeds. Modern large-scale text-to-image architectures heavily rely on computationally expensive guidance mechanisms to enforce complex prompt adherence. Mapping the bifurcation windows of massive latent diffusion and flow-based models \cite{song2023consistency,flux2024,labs2025flux1kontextflowmatching,flux-2-2025} could unlock highly optimized dynamic decoding algorithms. By anticipating exactly when a network binds independent semantic concepts, we can systematically accelerate inference pipelines without sacrificing the structural integrity of multi-concept synthesis.

%% file: ack.tex
\section*{Acknowledgement} 

This research is funded by Vietnam - Korea University of Information and Communication Technology under project number \emph{ÐHVH-2026-21}.